\documentclass[conference]{IEEEtran}
\IEEEoverridecommandlockouts
\usepackage{cite}
\usepackage{amsmath,amssymb,amsfonts}
\usepackage{algorithmic}
\usepackage{graphicx}
\usepackage{textcomp}
\usepackage{xcolor}
\usepackage{tikz}

\graphicspath{{images/}}
\DeclareGraphicsExtensions{.pdf,.jpeg,.jpg,.png}
\usepackage{mathptmx} 
\usepackage{times} 
\usepackage{amsmath} 
\usepackage{amssymb}  
\usepackage{hyperref}
\usepackage{tikz}

\usepackage{tabularx}
\usepackage{booktabs}

\def\BibTeX{{\rm B\kern-.05em{\sc i\kern-.025em b}\kern-.08em
    T\kern-.1667em\lower.7ex\hbox{E}\kern-.125emX}}

\begin{document}
\title{UAVs and Neural Networks for search and rescue missions}

\author{
\IEEEauthorblockN{1\textsuperscript{st} Hartmut Surmann}
\IEEEauthorblockA{\textit{Department of Computer Science} \\
\textit{Westphalian University}\\ \textit{of Applied Sciences}\\
Gelsenkirchen, Germany \\
hartmut.surmann@w-hs.de}
\and
\IEEEauthorblockN{2\textsuperscript{nd} Artur Leinweber}
\IEEEauthorblockA{\textit{Department of Computer Science} \\
\textit{Westphalian University}\\ \textit{of Applied Sciences}\\
Gelsenkirchen, Germany \\
artur.leinweber@w-hs.de}
\and
\IEEEauthorblockN{3\textsuperscript{rd} Gerhard Senkowski}
\IEEEauthorblockA{\textit{Department of Computer Science} \\
\textit{Westphalian University}\\ \textit{of Applied Sciences}\\
Gelsenkirchen, Germany \\
gerhard.senkowski@w-hs.de}
\vspace{0.3cm}
\and
\IEEEauthorblockN{\hspace{3.8cm}4\textsuperscript{th} Julien Meine}
\IEEEauthorblockA{\hspace{3.8cm}\textit{Department of Computer Science} \\
\hspace{3.8cm}\textit{Westphalian University}\\ \hspace{3.8cm}\textit{of Applied Sciences}\\
\hspace{3.8cm}Gelsenkirchen, Germany \\
\hspace{3.8cm}julien.meine@w-hs.de}
\and
\IEEEauthorblockN{5\textsuperscript{th} Dominik Slomma}
\IEEEauthorblockA{
\textit{German Rescue Robotic Centre}\\
Dortmund, Germany \\
dominik.10041991@web.de}
}

\maketitle

\begin{abstract}
In this paper, we present a method for detecting objects of interest, including cars, humans, and fire, in aerial images captured by unmanned aerial vehicles (UAVs) usually during vegetation fires. To achieve this, we use artificial neural networks and create a dataset for supervised learning. We accomplish the assisted labeling of the dataset through the implementation of an object detection pipeline that combines classic image processing techniques with pretrained neural networks. In addition, we develop a data augmentation pipeline to augment the dataset with automatically labeled images. Finally, we evaluate the performance of different neural networks.
\end{abstract}

\begin{IEEEkeywords}
SAR; UAV; AI; Object Detection; Vegetation Fires
\end{IEEEkeywords}

\section{Introduction}
\label{sec:introduction}
According to the European Forest Fire Information System (EFFIS), the frequency of vegetation fires such as forest fires increased sharply in recent years.
Depending on the extent of the fire, the effects are devastating. For example, ecosystems are destroyed, which affects the habitats of animals and humans. Additionally, pollutants are released into the environment during the combustion process, which can lead to health problems. While climate change is increasingly leading to situations that promote the development of vegetation fires, most fires are started by human activity. During fire fighting, the firefighters are exposed to the risk of being injured or killed. Mobile robots such as unmanned aerial vehicles can reduce the risks and support the emergency services during an operation so that they can, for example, gain an overview of the situation and initiate further fire-fighting measures\cite{9597869}\cite{9738529}.
The evaluation and analysis of aerial images, in order to identify relevant objects during vegetation fires takes a great deal of time and is usually carried out by one or more specialists. Since this is a time-critical task in search and rescue operations, machine vision can be used to detect objects in the aerial imagery to provide situational awareness in a timely manner. In this way, significant objects such as humans, vehicles and fire can be detected in the aerial images autmatically and the result is evaluated by a specialist and forwarded to the emergency services. For object detection, deep-learning methods have prevailed over other methods in research, since they are superior to those methods in terms of speed and accuracy \cite{DBLP:journals/corr/abs-1910-13796}. In most publications, new neural networks are presented and compared with existing ones\cite{forestfiredetectionzhang}\cite{deepconvneurnetfiredet}\cite{MUHAMMAD201830}\cite{8385121}\cite{8307064}. Since these are usually based on different architectures, each has its advantages and disadvantages, which means that the results differ. Usually, only one neural network is used in the context of search and rescue missions, although the rest of the methods also have potential to positively influence the results. Theoretically, there would be the possibility to merge the results by using different models. By adding more filters, the results would be more meaningful than using only one model.
Supervised learning of neural networks requires datasets that are authoritative for diverse scenarios. Acquiring data form real images, simulations, and other models such as Generative Adversarial Networks (GANs), is a time-consuming task. Additionally - if not already available - the objects in the aerial images usually have to be annotated with bounding boxes. In this context, different methods can be used to generate datasets through data augmentation with a lower expenditure of time, which can be used in supervised learning under certain assumptions. For example, it would be conceivable that objects that are recognized as false positives (FPs) serve as backgrounds. Objects that the model recognizes well or false negatives (FNs) can be placed in these backgrounds to increase the accuracy during training.
%

\section{Related Work}
Moffatt et. al. present a recent example in the field of fire detection \cite{9476715}. A hexacopter was equipped with a Velodyne VLP-16 for obstacle avoidance and fire detection. The algorithms for these tasks were run in real time on an Intel NUC onboard computer. For firefighting, a small fire extinguisher was developed and mounted under the UAV. The images from the infrared camera FLIR TAU were processed in two steps to detect fire. In the first step, darker and lighter areas in the image were determined through Local Intensity Operation (LIO) or Intensity Brightening Operation (IBO). In the second step, background noise was filtered out and a binary image was generated to show only the heat source. Based on this, the UAV can navigate to a fire source and extinguish it remotely by localizing the UAV using GPS and detecting the fire in the infrared image. This allows for the selective combat of individual sources of fire. Other handcrafted fire detection methods use a combination of techniques such as background subtraction, color spaces and spatial attributes to identify potential fire regions\cite{compvibameth}\cite{6636667}. With the advancement of deep learning techniques in Image Processing and Computer Vision, deep learning models have outperformed traditional handcrafted visual detection approaches\cite{deeplearninglecun}. Zhang et al. presented a deep learning approach for forest fire detection, which involves training a full image and patch-level classifier in a combined CNN. Their method has a cascaded approach, first using a global image-level classifier and then a fine-grained patch classifier to determine the exact location of fire\cite{forestfiredetectionzhang}. Samarth et al.\cite{DBLP:journals/corr/abs-1911-09010} proposed the use of CNNs for binary fire detection and superpixel localization in video or still imagery. They evaluated different reduced complexity CNN architectures including different Inception architectures, ResNet, and EfficientNet. They proposed two low-complexity variants of the InceptionV3 and InceptionV4 networks, InceptionV3-OnFire and InceptionV4-OnFire, as the best models for fire detection and localization. Others propose a multi-scale fire detection method using deep-stacked layers and a densely connected residual block. The final detection is made by considering predictions from different scales of feature maps through a weighted voting algorithm\cite{multiscalefire}. 

In this paper, we propose the use of multiple neural networks in the context of search and rescue missions, with the goal of improving the results compared to using only one network. Our approach merges the results of different models, incorporating more filters to achieve a more meaningful outcome. We understand the importance of having authoritative datasets for the supervised learning of these neural networks. Thus, we acquire data from real images, simulations, and other models such as Generative Adversarial Networks (GANs). However, obtaining this data can be time-consuming and may require manual annotation of objects in aerial images. To address this, we also examined methods for semi-automatic labeling of our data and ways to generate datasets with a lower expenditure of time for use in supervised learning.

\section{Data acquisition}
\subsection{Third party and internal datasets}
Within the context of vegetation fires, the dataset's focus is on fire scenarios. Various existing datasets for fire detection in images were considered during dataset creation. Robin Cole cites over 14 datasets containing fire images\footnote{Fire detection from images: \url{https://github.com/robmarkcole/fire-detection-from-images}}, and Xu et. al. utilized several publicly available datasets for supervised learning \cite{f12020217}. However, most available data lacks the camera perspective of a flying UAV. To support aerial assistance for responders, the imagery must include diverse perspectives, heights, and resolutions for a robust model.

A new 33GB dataset meeting these criteria was created and published on Kaggle\footnote{Aerial Rescue Object Detection: \url{https://www.kaggle.com/datasets/julienmeine/rescue-object-detection}}. A tool was developed to facilitate dataset creation by searching various media platforms, downloading sources, and extracting individual frames from relevant video segments. The dataset consists of historical vegetation fires filmed from the ground or air, as well as internal data from real disaster areas, such as vegetation fires, industrial fires, and floods, or from SAR exercises like searches for people and vehicles. With about 21,000 third-party images and 5,000 internal images, approximately 10\% of the 26,000 total images were used for training, as most data lacked labels.

\subsection{Generative Adversarial Networks}

Generative Adversarial Networks (CycleGAN\cite{8237506}) were explored for acquiring data in different fire scenarios, as aerial photographs are not always available. CycleGAN was used to map data from one context to another, e.g., from an empty burning crop field to a crop field with a burning barn. Frames were extracted from two public UAV-captured videos, creating a dataset of 2588 images, with 200 for evaluation.

\begin{figure}[ht]
\centering
\includegraphics[width=\columnwidth]{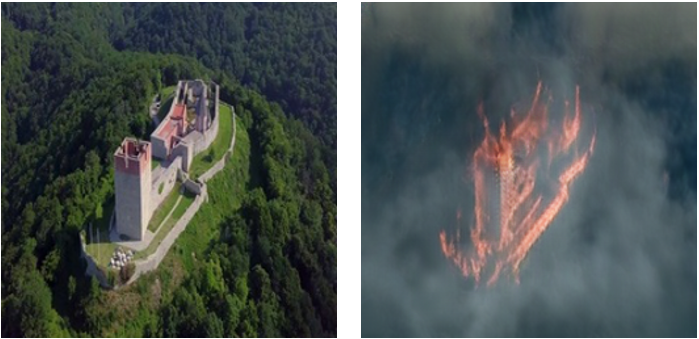}
\caption{Input images (left) and the fake results from CycleGAN (right).}
\label{fig:gan}
\end{figure}

Test results showed CycleGAN learning a scenario transfer, but generated images were unconvincing, inaccurately depicting fires and buildings (Figure \ref{fig:gan}). Consequently, none were included in the dataset. Despite this, the approach may be successful for other problems, given different data and hyperparameters.

\subsection{Simulation}
Simulations offer a means of creating scenarios and delivering information that closely resembles reality. However, such works often prioritize physical events, like the varying burn rates of different tree types, over realistic graphics \cite{10.1145/3450626.3459954}. As a result, a game engine was explored for additional data collection. Several projects within the "Unreal Engine 4" (UE4) were examined, ultimately selecting the prefabricated 'Landscape Mountains' project, which features a mountainous landscape with trees, a lake, and other objects.

\begin{figure}[ht]
\centering
\includegraphics[width=\columnwidth]{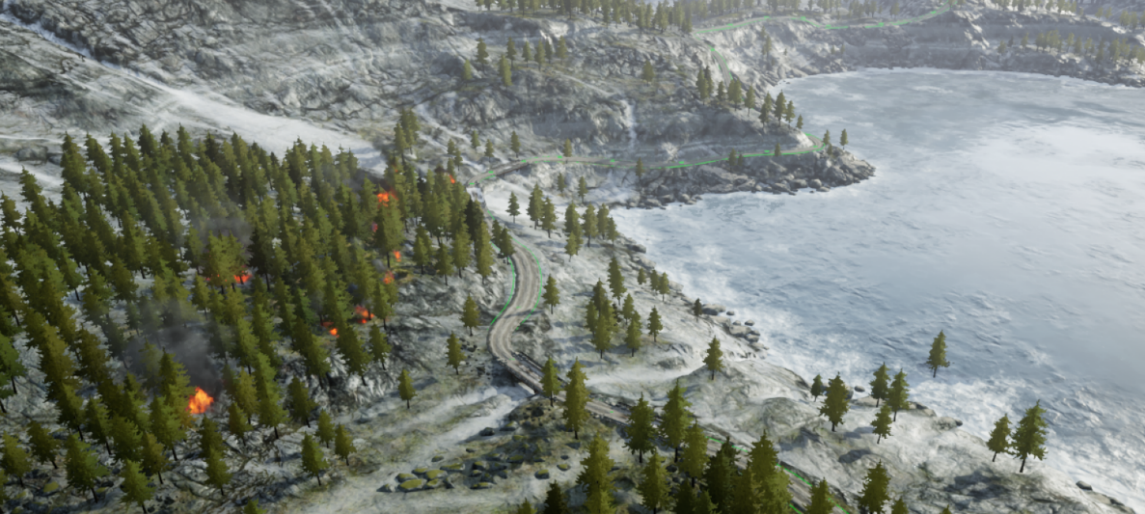}
\caption{Unreal Engine 4 'Landscape Mountains' project with flames simulate a forest fire.}
\label{fig:sim}
\end{figure}

Fire collections were incorporated into the project, positioned at different locations amid the trees, to create a visually realistic and interactive representation of a fire, rather than a physical image. Flames were placed near the ground and at tree height, simulating an early-stage forest fire with various flame sizes at different locations (Figure \ref{fig:sim}).

To generate aerial photographs, a script was developed to simulate a meandering UAV flight trajectory, producing around 150 aerial images from varying heights and perspectives at regular intervals.

\section{Dataset Labeling}
\begin{figure}[ht]
\centering
\includegraphics[width=\columnwidth]{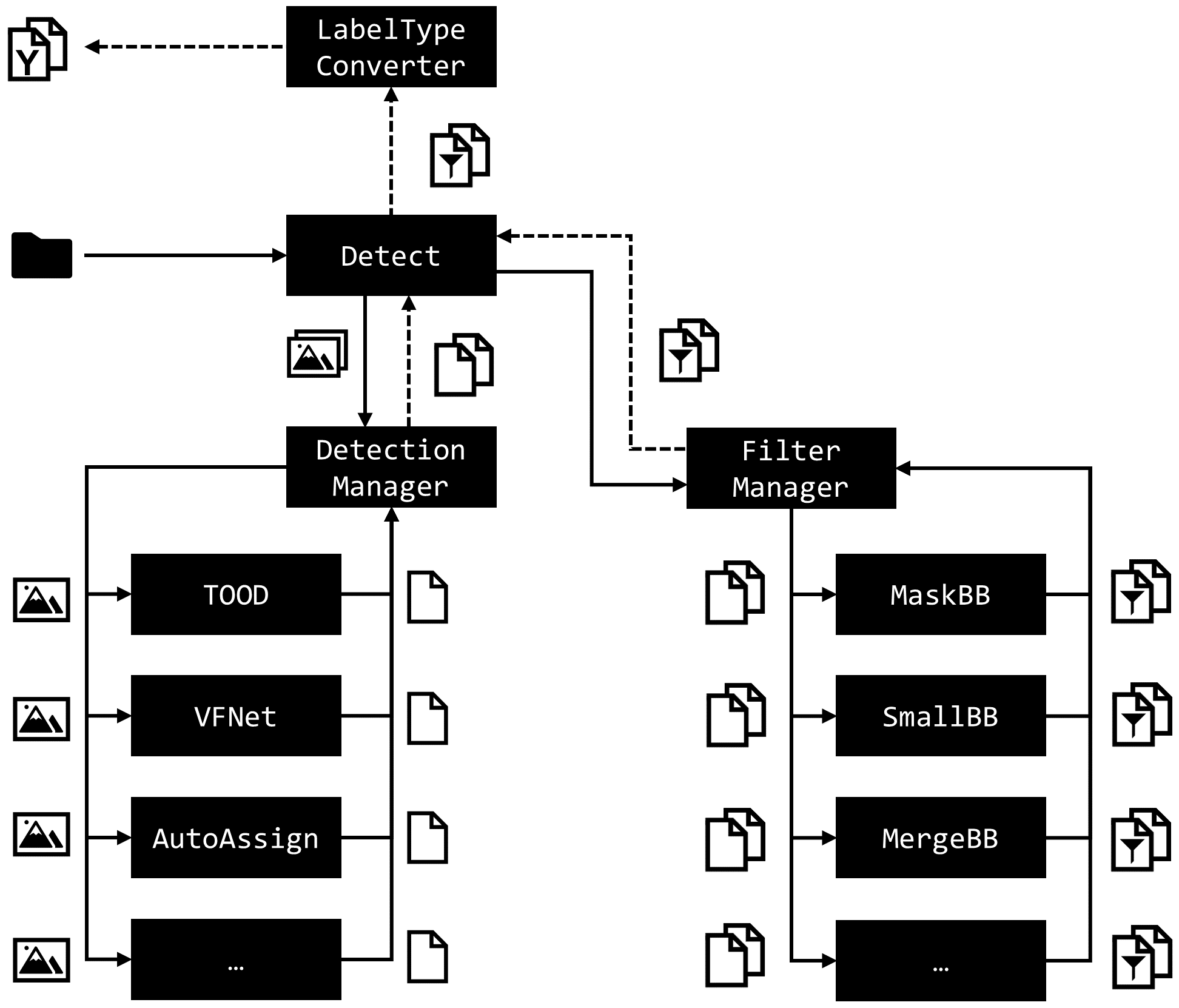}
\caption{Architecture of the Object Detection Pipeline. The 'DetectionManager' class manages the 'Detector' instances, e.g. 'TOOD', and tasks them with detecting objects in the images. After all images have been commissioned by the 'DetectionManager' and the 'Detectors' have processed each image, the results are passed to a 'FilterManager'. Similar to the 'DetectionManager', the 'FilterManager' manages different 'Filter' instances where the results are fused or filtered. After filtering, the bounding boxes are converted into another convention of an annotation by means of the class 'LabelTypeConverter'.}
\label{fig:odp}
\end{figure}

There are various types of services and methods for labeling bounding boxes in images. One possibility is Model-Assisted Labeling (MAL), where data is labeled with the help of neural networks. This basic principle was adopted and implemented for this work. Since not only neural networks but also conventional computer vision algorithms were used in this case, the method is referred to as assisted labeling (AL) in this paper. The following pretrained object localization and recognition methods were used:

\begin{itemize}
    \item Rule-Based Color-Model (Fire)
    \item Faster R-CNN (Vehicle)
    \item YOLOv3 (Fire, Vehicle)
    \item Light-Weight RefineNet (Human)
\end{itemize}
Using these methods, 2400 images were processed by the Object Detection Pipeline within 3 hours. The Object Detection Pipeline (ODP) is primarily used for the integration of several object detection methods, whose results are subsequently filtered and merged. The reason for building this pipeline (Figure \ref{fig:odp}) is mainly based on the following aspect: The competition between authors and developers in publications about object detection methods, especially neural networks, is based on the development of different approaches and architectures whose results are finally compared. These object detection architectures often take different approaches and thus have opposing strengths and weaknesses, the combination of which can lead to positive results. Thus, deficits of one method can be compensated for with the strenght of another method, for example, deficits in the detection of small objects of one method can be compensated for with another method whose strength is the detection of small objects. 

\begin{figure}[ht]
\centering
\includegraphics[width=\columnwidth]{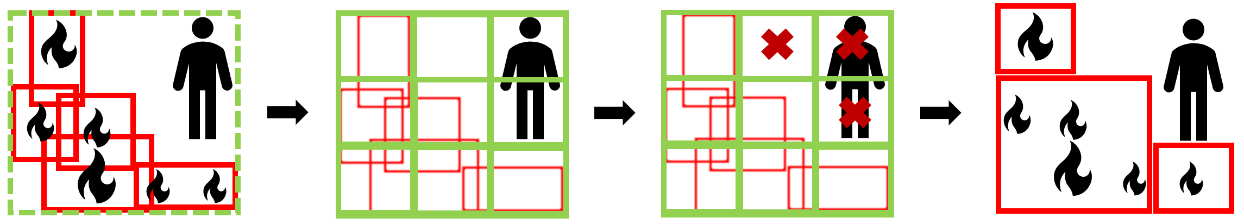}
\caption{Sequence of the four steps of the 'MaskBB' filter, where five bounding boxes were combined to three bounding boxes.}
\label{fig:filter}
\end{figure}

All images are processed by the various neural networks, and the resulting bounding boxes are passed to the filter manager for further processing. We have implemented several filters to modify the bounding boxes. The 'SmallBB' filter removes bounding boxes that are smaller than a certain threshold in width and height. The 'MaskBB' filter (shown in Figure \ref{fig:filter}) first enlarges all bounding boxes by a factor, combines bounding boxes of the same class that overlap, and creates a large bounding box. In the second step, the large bounding box is split into nine smaller bounding boxes to reduce background and exclude objects that do not belong to the class. In the third step, bounding boxes that do not overlap with the original bounding boxes are removed. Finally, the remaining bounding boxes are merged to form the smallest number of large bounding boxes possible. The 'MergeBB' filter merges bounding boxes of the same class that have an intersection over union (IoU) or generalized intersection over union (GIoU) greater than a certain threshold.

The test of the filter was completed with 200 randomly selected images of the dataset. Before filtering, the 200 images contained about 6000 bounding boxes. Filtering with the ODP reduced the number to about 500 bounding boxes using several filters. After manual correction the number increased to approximately 600 bounding boxes. This corresponds to a decrease of approx. 90\%. Based on this, the remaining bounding boxes in the dataset were manually corrected. Compared to ODP, manual labeling required ten times the time in terms of correction. Furthermore, concentration plays a significant role in manual labeling, so it cannot be ruled out that errors occur during labeling. For this reason, several cycles are run during AL, depending on the time budget, in order to keep the number of errors low.

\section{First learning cycle}
\begin{figure}[ht]
    \centering
    \begin{tikzpicture}
        \filldraw[fill=blue!30,draw=blue!50!black] (0,0.8) rectangle (6.34,1.4) node[midway,right] {6340};
        \filldraw[fill=blue!30,draw=blue!50!black] (0,1.6) rectangle (4.35,2.2) node[midway,right] {4354};
        \filldraw[fill=blue!30,draw=blue!50!black] (0,2.4) rectangle (6.62,3.0) node[midway,right] {6622};
        \filldraw[fill=red!30,draw=red!50!black] (0,0.8) rectangle (0.64,1.4) node[pos=0.4,pos=0.5] {640};
        \filldraw[fill=red!30,draw=red!50!black] (0,1.6) rectangle (0.56,2.2) node[pos=0.4,pos=0.5] {563};
        \filldraw[fill=red!30,draw=red!50!black] (0,2.4) rectangle (0.68,3.0) node[pos=0.4,pos=0.5] {682};
        \node[anchor=south] at (3,3.2) {\textbf{Bounding box distribution}};
        \node[anchor=east] at (-0.2,1.1) {\textbf{Fire}};
        \node[anchor=east] at (-0.2,1.9) {\textbf{Vehicle}};
        \node[anchor=east] at (-0.2,2.7) {\textbf{Human}};
    \end{tikzpicture}
    \caption{Distribution of the 19201 bounding boxes for testing(red) and training(blue)}
    \label{fig:flc}
\end{figure}
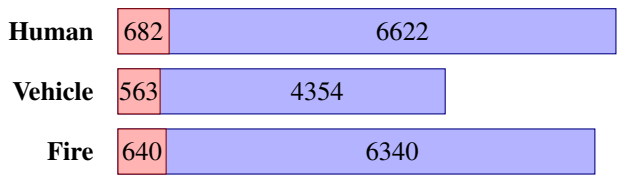

After labeling, the dataset was split into a training dataset of 2334 images and an evaluation dataset of 259 images, resulting in a 90:10 ratio with over 19000 bounding boxes (see figure \ref{fig:flc}).

We used OpenMMLab's MMDetection toolbox for training. Exemplarily, the following neural networks were selected. These were available in the framework at the time of the investigation and could demonstrate good metrics on the COCO dataset. The YOLOX network performs slightly worse in comparison, but it is a fast network and was therefore considered in the context of search and rescue.

{\small
\begin{itemize}
    \item TOOD (Version: tood r101 fpn dconv c3-c5 mstrain 2x)
    \item AutoAssign (Version: autoassign r50 fpn 8x2 1x)
    \item YOLOX (Version: yolox s 8x8 300e)
    \item VarifocalNet (Version: vfnet x101 32x4d fpn mdconv c3-c5 mstrain 2x)
    \item Deformable DETR (Version: deformable detr twostage refine r50 16x2 50e)
\end{itemize}
}




For training, we downloaded the weights of the pre-trained models that resulted from the supervised learning of the COCO dataset.

Table \ref{tab:valsl1} shows that during training, the mAP of the Deformable DETR is worse for small objects than for the AutoAssign, which in turn does not perform as well as the Deformable DETR for objects of medium size. In all three categories, the mAPs for the YOLOX are the lowest, whereas the TOOD performs best.
\begin{table}[ht]
\centering
\caption{mAPs and mARs from the first learning cycle}
\label{tab:valsl1}
\centering
\begin{tabular}{p{1cm}p{1cm}p{1cm}p{1cm}p{1cm}p{1cm}}
\cmidrule[\heavyrulewidth](lr){1-6}
\textbf{mAP}
& \begin{tabular}[c]{@{}l@{}}\textbf{TOOD}\\ \textbf{Epoch 71}\end{tabular} 
& \begin{tabular}[c]{@{}l@{}}\textbf{AutoAssign}\\ \textbf{Epoch 71}\end{tabular} 
& \begin{tabular}[c]{@{}l@{}}\textbf{YOLOX}\\ \textbf{Epoch 100}\end{tabular} 
& \begin{tabular}[c]{@{}l@{}}\textbf{VFNet}\\ \textbf{Epoch 80}\end{tabular} 
& \begin{tabular}[c]{@{}l@{}}\textbf{DETR}\\\textbf{Epoch 98}\end{tabular} \\
\cmidrule(lr){1-6}
\begin{tabular}[c]{@{}l@{}}IoU=\\0.50:0.95\end{tabular} & 0.380 & 0.341 & 0.302 & 0.340 & 0.362 \\
\begin{tabular}[c]{@{}l@{}}IoU=0.50\end{tabular} & 0.745 & 0.700 & 0.644 & 0.678 & 0.738 \\
\begin{tabular}[c]{@{}l@{}}IoU=0.75\end{tabular} & 0.349 & 0.295 & 0.238 & 0.296 & 0.308 \\
\begin{tabular}[c]{@{}l@{}}small\end{tabular} & 0.176 & 0.159 & 0.110 & 0.115 & 0.151 \\
\begin{tabular}[c]{@{}l@{}}medium\end{tabular} & 0.350 & 0.317 & 0.283 & 0.311 & 0.343 \\
\begin{tabular}[c]{@{}l@{}}large\end{tabular} & 0.474 & 0.447 & 0.392 & 0.445 & 0.456 \\
\begin{tabular}[c]{@{}l@{}}mAR @ \\IoU=\\0.50:0.95\end{tabular} & 0.506 & 0.483 & 0.445 & 0.477 & 0.494\\
\cmidrule[\heavyrulewidth](lr){1-6}
\end{tabular}
\end{table}

The goal of this learning cycle was to identify the strengths and weaknesses of each model. To do so, we incorporated the models into the object detection pipeline, loaded the dataset, and extracted the FPs and FNs. In this context, a bounding box was considered a FN if its IoU with the ground truth was zero. This criterion highlights bounding boxes that were not detected by the neural networks as FNs and reduces the number of FPs. A confidence score of 0.3 was used.
Tables \ref{tab:fps} and \ref{tab:fns} show the number of FPs and FNs for each class of neural network. The AutoAssign method followed by the Deformable DETR have the lowest number of FPs, while the YOLOX and VFNet have the highest. The TOOD falls in the middle. The results for the FNs are similar to the FPs: the Deformable DETR followed by the AutoAssign have a low number of FNs, while the YOLOX performs the worst. However, the TOOD generates a similar number of FNs as the YOLOX. This suggests that while the TOOD has the highest mean average precision (mAP) and mean average recall (mAR), making it the overall best method, it has more difficulty than the AutoAssign and Deformable DETR in identifying specific objects in the images.

\begin{table}[ht]
\centering
\caption{FPs of the models from the first learning cycle}
\label{tab:fps}
\centering
\begin{tabular}{ccccc}
\cmidrule[\heavyrulewidth](lr){1-5}
\textbf{Model} & \textbf{Fire} & \textbf{Vehicle} & \textbf{Human} & \textbf{Sum} \\
\cmidrule(lr){1-5}
TOOD            & 17   & 86      & 69    & 172   \\
AutoAssign      & 2    & 2       & 8     & 12    \\
YOLOX           & 3    & 208     & 142   & 353   \\
VFNet           & 20   & 127     & 141   & 288   \\
Deformable DETR & 11   & 38      & 63    & 112  \\
\cmidrule[\heavyrulewidth](lr){1-5}
\end{tabular}
\end{table}

\begin{table}[ht]
\centering
\caption{FNs of the models from the first cycle of learning}
\label{tab:fns}
\centering
\begin{tabular}{ccccc}
\cmidrule[\heavyrulewidth](lr){1-5}
\textbf{Model} & \textbf{Fire} & \textbf{Vehicle} & \textbf{Human} & \textbf{Sum} \\
\cmidrule(lr){1-5}
TOOD            & 53   & 39      & 170   & 262   \\
AutoAssign      & 9    & 18      & 21    & 48    \\
YOLOX           & 71   & 162     & 234   & 467   \\
VFNet           & 29   & 30      & 156   & 215   \\
Deformable DETR & 8    & 12      & 20    & 40   \\
\cmidrule[\heavyrulewidth](lr){1-5}
\end{tabular}
\end{table}

Some objects were not detected during the dataset labeling process, but were subsequently detected by the neural networks. This enabled us to make corrections. After the corrections were made, the FPs and FNs from the five neural networks were merged using the ODP and the MergeBB filter, which combines bounding boxes of all classes that have an IoU of at least 0.3. This led to a decrease in the number of FNs, as the detection coverage in this case was provided by neural networks such as the Deformable DETR. In the process, 358 relevant objects were detected but were incorrectly classified and counted as FPs. To obtain only irrelevant FPs, we manually reviewed the results and removed these objects (see Table \ref{tab:fpsfnscorrect}).

\begin{table}[ht]
\caption{Number of FPs and FNs after manual correction}
\label{tab:fpsfnscorrect}
\centering
\begin{tabular}{cccc}
\cmidrule[\heavyrulewidth](lr){1-4}
   & \textbf{Fire} & \textbf{Vehicle} & \textbf{Human} \\
\cmidrule(lr){1-4}
FP & 224  & 621     & 880   \\
FN & 129  & 104     & 20   \\
\cmidrule[\heavyrulewidth](lr){1-4}
\end{tabular}
\end{table}

\section{Mosaic-Augmentation}
\begin{figure}[ht]
\centering
\includegraphics[width=\columnwidth]{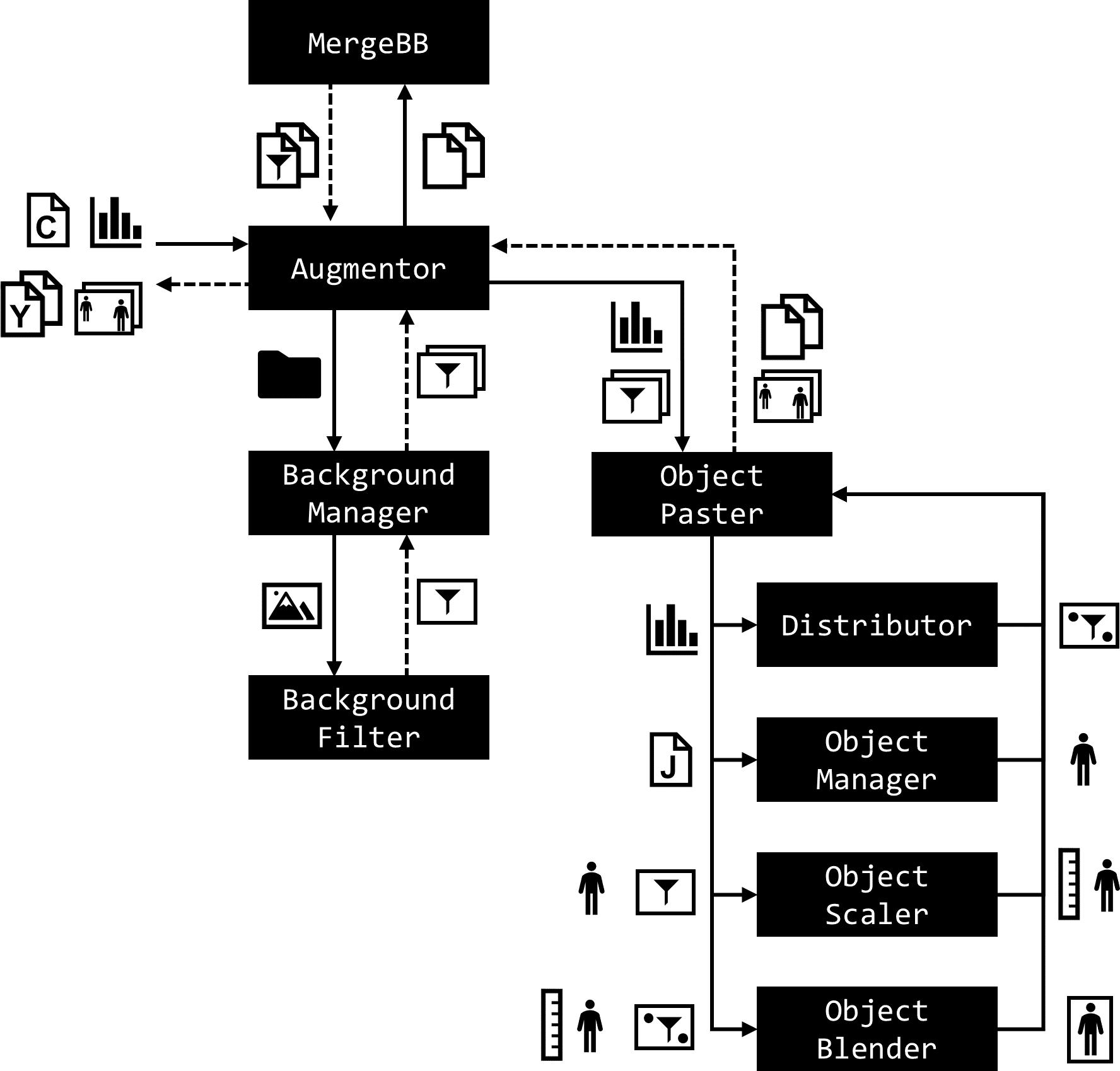}
\caption{The Data Augmentation Pipeline (DAP) consists of several components. The 'BackgroundManager' manages 'Background' instances and allows the backgrounds to be modified using 'BackgroundFilter' image processing algorithms. 'ObjectClass' represents a relevant object and includes an attribute specifying the maximum width of the object in meters. The relevant objects are managed by an 'ObjectManager', and the 'ObjectPaster' is responsible for inserting the objects into the backgrounds. The 'Distributor' calculates the position and number of objects of each class to be inserted. The 'ObjectScaler' can scale the objects and backgrounds to a realistic size based on their EXIF header and metric attribute, so as not to lose context in certain situations. The 'ObjectBlender' class applies image processing algorithms to blend the objects into the background and adjust the contrast, brightness, and color to match the background where the objects overlap. This changes the visual appearance of the objects depending on the background.}
\label{fig:dap}
\end{figure}

\begin{figure}[ht]
\centering
\includegraphics[width=\columnwidth]{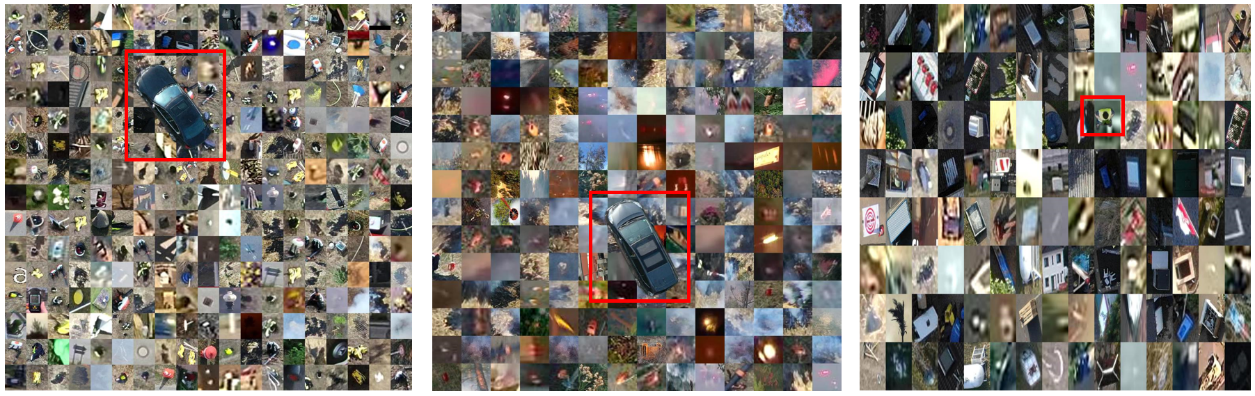}
\caption{Mosaic augmentation using the DAP from Human FPs (left), Fire FPs (center), and Vehicle FPs (right).}
\label{fig:mosaic}
\end{figure}

After training, the results (e.g., FPs and FNs) are often not further considered. The Data Augmentation Pipeline (DAP) addresses this issue (see Figure \ref{fig:dap}). The DAP uses the ODP to extract the FPs and FNs, as well as backgrounds that do not contain relevant objects. With this information, the DAP creates a new dataset that projects known objects onto new backgrounds to help the neural networks distinguish relevant objects from irrelevant objects and separate them from the backgrounds during a renewed training.
An experiment was conducted to increase the mAP and mAR after a re-learning run. The FPs were grouped into mosaic-like images using a tool of the DAP. In this process, another aspect was considered: the grouping of the classes from which the FPs originated. This means that FPs that were misclassified as flames, for example, were assembled exclusively with other FPs that were also misclassified as flames. The mosaic-like images were generated using three different variants. In the first variant, the resolution of an FP was scaled quadratically. In the last two variants, the width and height of the resolution of an FP were doubled, respectively.
During the creation of a mosaic-like image, the FPs of a class were randomly shuffled and then assembled. The number of iterations was determined by the total number of FPs and the number needed for a mosaic-like image. In addition to the minimum number of iterations required, nine more iterations were performed. This ultimately generated over 300 images (Fire: 50; Human: 140; Vehicle: 120) with a resolution of approximately 900 × 900 pixels. Since the generated images did not contain relevant objects, they could not be immediately used for supervised learning. Therefore, relevant objects or bounding boxes were extracted from aerial photographs taken in a bird's eye view from internal datasets using the object detection pipeline. To insert objects into the generated mosaic-like images, the background of the objects was manually masked, resulting in a total of over 100 objects (Fire:17; Human:25; Vehicle:61). This process was facilitated by the DAP (see Figure \ref{fig:mosaic}). Only one randomly chosen object was added to a randomly chosen coordinate in each image. Care was taken to ensure that the class of objects added to the mosaic-like images did not match. This means that a mosaic-like image created from FPs of flames did not have an object of class 'flame' added to it, but rather an object of class 'vehicle', for example. Using an opposite class for the object creates a contrast with the background of the FPs within the bounding box.

\section{Object-Pasting}
\begin{figure}[ht]
\centering
\includegraphics[width=\columnwidth]{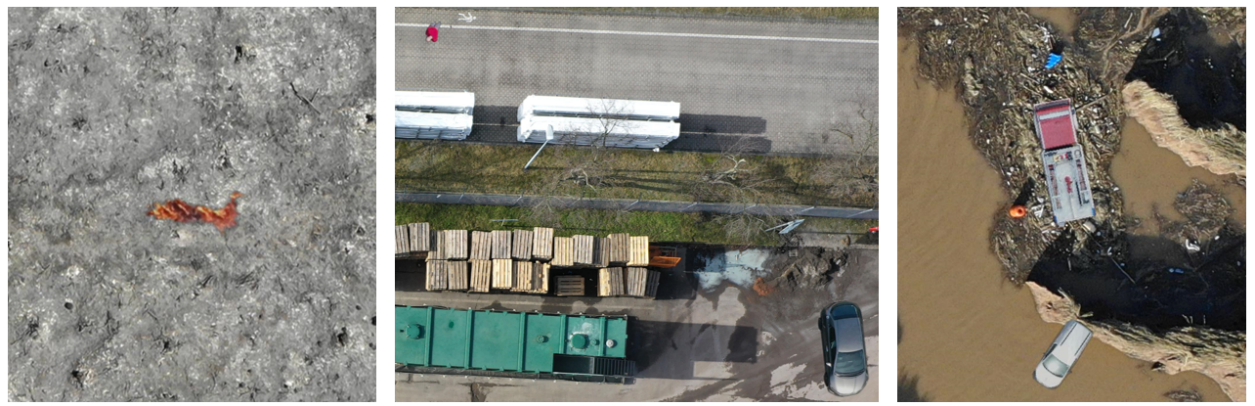}
\caption{Object pasting using the Data Augmentation Pipeline. Examples include a flame on a dry vegetation ground (left), a person and a vehicle in an industrial area (center), and two vehicles in a flooded area (right).}
\label{fig:pasting}
\end{figure}

Aerial images that did not contain any relevant objects were processed with the DAP. As a result, over 190 aerial photographs were automatically labeled with a total of over 1000 objects (fire:96, vehicle: 573, human: 423). The EXIF data allowed the objects to be scaled realistically based on the flight altitude. Different configurations were used to insert the objects into each dataset. Figure \ref{fig:pasting} shows example results of the DAP. If the EXIF headers of the images contained more information, it would be possible to add shadows to people and vehicles or simulate smoke for flames. However, the context of the images cannot be considered, so an added vehicle on the roof would be the same size as an added vehicle on the road. Despite this, the aerial photographs contain background information that can be useful for the neural networks in supervised learning to optimize the FPs and FNs by differentiating the objects from the backgrounds.

\section{Second learning cycle}
After adding the images to the dataset, a total of 3097 images with 21454 bounding boxes were available for the next learning cycle. The evaluation dataset was not modified so that the learning cycles could be compared retrospectively. In this cycle, the number of epochs for the YOLOX was increased to 300, because the YOLOX always performed best at epoch 100 in the second learning cycle, indicating that its full potential was not being utilized. 

The remaining hyperparameters of the configuration files were not changed. For the training, the original weights of the pre-trained neural networks that were created during the supervised learning of the COCO dataset were used again to avoid the risk of getting stuck in a local optimum. The results of the same models from the first learning cycle are compared with the respective models from the second learning cycle in Table \ref{tab:sl2val}.

\begin{table}[ht]
\centering
\caption{mAPs and mARs in comparison from the first and second learning cycle}
\label{tab:sl2val}
\centering
\scalebox{0.86}{
\begin{tabular}{p{0.95cm}p{0.8cm}p{0.8cm}p{0.8cm}p{0.8cm}p{0.8cm}p{0.8cm}p{1cm}}
\cmidrule[\heavyrulewidth](lr){1-8}
\begin{tabular}[c]{@{}l@{}}\textbf{Model}\\\textbf{Epoch}\end{tabular} 
& \begin{tabular}[c]{@{}l@{}}\textbf{mAP @}\\\textbf{IoU=}\\\textbf{0.50:0.95}\end{tabular} 
& \begin{tabular}[c]{@{}l@{}}\textbf{mAP @}\\\textbf{IoU=}\\\textbf{0.50}\end{tabular} 
& \begin{tabular}[c]{@{}l@{}}\textbf{mAP @}\\\textbf{IoU=}\\\textbf{0.75}\end{tabular} 
& \begin{tabular}[c]{@{}l@{}}\textbf{mAP}\\\textbf{small}\end{tabular} 
& \begin{tabular}[c]{@{}l@{}}\textbf{mAP}\\\textbf{medium}\end{tabular} 
& \begin{tabular}[c]{@{}l@{}}\textbf{mAP}\\\textbf{large}\end{tabular} 
& \begin{tabular}[c]{@{}l@{}}\textbf{mAR @}\\\textbf{IoU=}\\\textbf{0.50:0.95}\end{tabular} \\
\cmidrule(lr){1-8}
\begin{tabular}[c]{@{}l@{}}1. TOOD\end{tabular} & 0.380 & 0.745 & 0.349 & 0.176 & 0.350 & 0.474 & 0.506 \\
\begin{tabular}[c]{@{}l@{}}2. TOOD\end{tabular} & 0.383 & 0.749 & 0.342 & 0.187 & 0.360 & 0.481 & 0.524 \\
\begin{tabular}[c]{@{}l@{}}1. AutoA\end{tabular} & 0.341 & 0.700 & 0.295 & 0.159 & 0.317 & 0.447 & 0.483 \\
\begin{tabular}[c]{@{}l@{}}2. AutoA\end{tabular} & 0.352 & 0.712 & 0.311 & 0.156 & 0.326 & 0.456 & 0.511 \\
\begin{tabular}[c]{@{}l@{}}1. YOLOX\end{tabular}& 0.302 & 0.644 & 0.238 & 0.110 & 0.283 & 0.392 & 0.445\\
\begin{tabular}[c]{@{}l@{}}2. YOLOX\end{tabular}& 0.352 & 0.701 & 0.306 & 0.137 & 0.319 & 0.454 & 0.460\\
\begin{tabular}[c]{@{}l@{}}1. VFNet\end{tabular} & 0.340 & 0.678 & 0.296 & 0.115 & 0.311 & 0.445 & 0.477 \\
\begin{tabular}[c]{@{}l@{}}2. VFNet\end{tabular} & 0.350 & 0.692 & 0.302 & 0.135 & 0.313 & 0.461 & 0.475 \\
\begin{tabular}[c]{@{}l@{}}1. DDETR\end{tabular}& 0.362 & 0.738 & 0.308 & 0.151 & 0.343 & 0.456 & 0.494 \\
\begin{tabular}[c]{@{}l@{}}2. DDETR\end{tabular} & 0.370 & 0.744 & 0.309 & 0.177 & 0.356 & 0.460 & 0.500 \\
\cmidrule[\heavyrulewidth](lr){1-8}
\end{tabular}
}
\end{table}

Table \ref{tab:sl2val} compares the top five models in the second learning cycle.

\begin{table}[ht]
\caption{mAPs and mARs from the second learning cycle}
\label{tab:sl2val2}
\centering
\scalebox{0.95}{
\begin{tabular}{p{1cm}p{1cm}p{1cm}p{1cm}p{1cm}p{1.5cm}}
\cmidrule[\heavyrulewidth](lr){1-6}
\textbf{mAP}
& \begin{tabular}[c]{@{}l@{}}\textbf{TOOD}\end{tabular}
& \begin{tabular}[c]{@{}l@{}}\textbf{AutoAssign}\end{tabular}
& \begin{tabular}[c]{@{}l@{}}\textbf{YOLOX}\end{tabular} 
& \begin{tabular}[c]{@{}l@{}}\textbf{VFNet}\end{tabular} 
& \begin{tabular}[c]{@{}l@{}}\textbf{Deformable} \\\textbf{DETR}\end{tabular} \\
\cmidrule(lr){1-6}
\begin{tabular}[c]{@{}l@{}}IoU=\\0.50:0.95\end{tabular} & 0.383 & 0.352 & 0.352 & 0.350 & 0.370 \\
\begin{tabular}[c]{@{}l@{}}IoU=0.50\end{tabular} & 0.749 & 0.712 & 0.701 & 0.692 & 0.744 \\
\begin{tabular}[c]{@{}l@{}}IoU=0.75\end{tabular} & 0.342 & 0.311 & 0.306 & 0.302 & 0.309 \\
\begin{tabular}[c]{@{}l@{}}small\end{tabular} & 0.187 & 0.156 & 0.137 & 0.135 & 0.177 \\
\begin{tabular}[c]{@{}l@{}}medium\end{tabular} & 0.360 & 0.326 & 0.319 & 0.313 & 0.356 \\
\begin{tabular}[c]{@{}l@{}}large\end{tabular} & 0.481 & 0.456 & 0.454 & 0.461 & 0.460 \\
\begin{tabular}[c]{@{}l@{}}mAR @ \\ IoU=\\ 0.50:0.95\end{tabular} & 0.524 & 0.511 & 0.460 & 0.475 & 0.500 \\
\cmidrule[\heavyrulewidth](lr){1-6}
\end{tabular}
}
\end{table}

The TOOD outperforms the other models in all metrics, followed by the Deformable DETR and the AutoAssign. The YOLOX and the VFNet show similar results. The following table shows the mAP @[IoU=0.50:0.95] for the respective classes, noting that the values are significantly higher for mAP @[IoU=0.50].

Table \ref{tab:sl2val3} shows that the TOOD performs well for the 'Fire' and 'Vehicle' classes, but the Deformable DETR performs better for the 'Human' class. To further evaluate the effectiveness of the mosaic augmentation and object pasting, row-normalized and column-normalized confusion matrices were computed from the first and second learning cycles for the respective models. The matrices were calculated using default values such as an IoU of 0.5 and a confidence score of 0.3. On average, the number of FPs increased by about 16\%, while the number of FNs decreased by about 10\%.
It is clear from the results of the experiment that the object pasting and mosaic augmentation techniques had a positive effect on the performance of the models in the second learning cycle. The mAP and mAR of the models increased, and the number of FNs decreased, while the number of FPs increased. This suggests that these techniques can be useful for improving the performance of neural networks in object detection tasks. It would be interesting to further explore the potential of these techniques in different contexts and with different neural network architectures, in order to more fully understand their capabilities and limitations. Additionally, it would be useful to investigate methods for minimizing the increase in FPs while still achieving the benefits of the object pasting and mosaic augmentation techniques.

\begin{figure}[ht]
    \centering
    \begin{tikzpicture}
        \filldraw[fill=blue!30,draw=blue!50!black] (0,0.8) rectangle (6.85,1.4) node[midway,right] {6854};
        \filldraw[fill=blue!30,draw=blue!50!black] (0,1.6) rectangle (5.43,2.2) node[midway,right] {5430};
        \filldraw[fill=blue!30,draw=blue!50!black] (0,2.4) rectangle (7.29,3.0) node[midway,right] {7285};
        \filldraw[fill=red!30,draw=red!50!black] (0,0.8) rectangle (0.64,1.4) node[pos=0.4,pos=0.5] {640};
        \filldraw[fill=red!30,draw=red!50!black] (0,1.6) rectangle (0.56,2.2) node[pos=0.4,pos=0.5] {563};
        \filldraw[fill=red!30,draw=red!50!black] (0,2.4) rectangle (0.68,3.0) node[pos=0.4,pos=0.5] {682};
        \node[anchor=south] at (3,3.2) {\textbf{Bounding box distribution}};
        \node[anchor=east] at (-0.2,1.1) {\textbf{Fire}};
        \node[anchor=east] at (-0.2,1.9) {\textbf{Vehicle}};
        \node[anchor=east] at (-0.2,2.7) {\textbf{Human}};
    \end{tikzpicture}
    \caption{Distribution of bounding boxes for testing(red) and training(blue)}
    \label{fig:sl2}
\end{figure}

\section{Proof of Concept}
To further evaluate the performance of the models after the second learning cycle, a test was conducted using 282 images of fire from synthetic data and from a vegetation fire exercise. These images were not used in the training or evaluation process, so the results are based on data that the models have not previously processed. The test used 154 images of synthetic data containing only fire and 128 images from an internal dataset of a vegetation fire exercise, totaling 2788 bounding boxes.


\begin{table}[ht]
\centering
\caption{mAPs @ IoU=0.50:0.95 of the models of the respective classes from the second learning cylce}
\label{tab:sl2val3}
\centering
\begin{tabular}{cccc}
\cmidrule[\heavyrulewidth](lr){1-4}
\textbf{Model}
& \textbf{Fire}
& \textbf{Vehicle}
& \textbf{Human} \\
\cmidrule(lr){1-4}
\begin{tabular}[c]{@{}l@{}}TOOD\end{tabular} & 0.229 & 0.535 & 0.385 \\
\begin{tabular}[c]{@{}l@{}}AutoAssign\end{tabular} & 0.228 & 0.482 & 0.347 \\
\begin{tabular}[c]{@{}l@{}}YOLOX\end{tabular} & 0.218 & 0.507 & 0.330 \\
\begin{tabular}[c]{@{}l@{}}VFNet\end{tabular} & 0.209 & 0.507 & 0.334 \\
\begin{tabular}[c]{@{}l@{}}Deformable DETR\end{tabular} & 0.208 & 0.509 & 0.392 \\
\cmidrule[\heavyrulewidth](lr){1-4}
\end{tabular}
\end{table}

\begin{figure}
\centering
\includegraphics[width=\columnwidth]{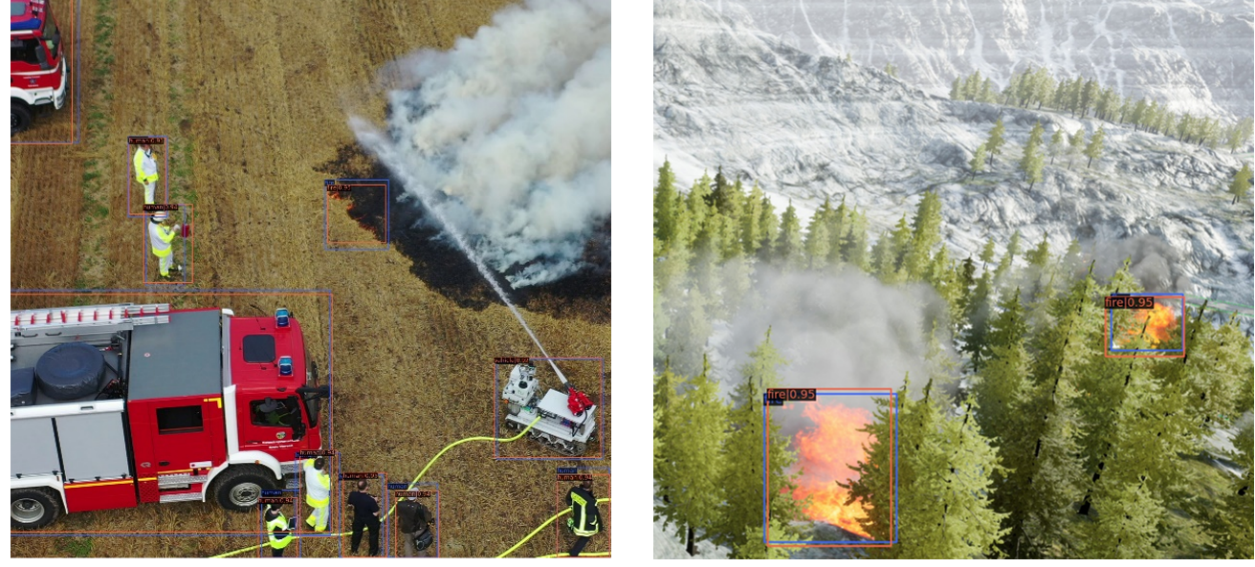}
\caption{Deformable-DETR inference with real aerial image (left) and artificial aerial image (right).}
\label{fig:t1}
\end{figure}

The results of the models on the test dataset show that the TOOD performs the best, followed by the Deformable DETR and the AutoAssign. The YOLOX and VFNet show similar performance. The results for the 'Vehicle' and 'Human' classes are similar to those on the evaluation dataset, but the models perform worse for the 'Fire' class on the test dataset. However, the models are still able to provide consistent results. Exemplary images from the test dataset processed by the neural networks are shown in figure \ref{fig:t1}.

\begin{figure}
\centering
\includegraphics[width=\columnwidth]{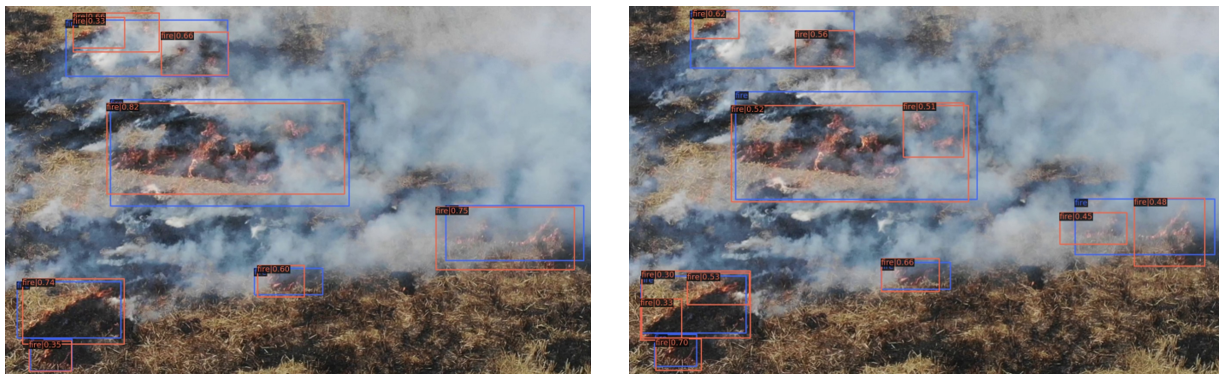}
\caption{Test image YOLOX (left) and TOOD (right).}
\label{fig:t3}
\end{figure}

The test dataset showed that the performance of the models varies depending on the granularity of the objects being detected, particularly for the 'Fire' class. One example is the difference between the results obtained by the TOOD and YOLOX models, where the TOOD had bounding boxes that did not match the ground truth bounding boxes in the lower right corner of an image, while the YOLOX's bounding boxes were almost identical to the ground truth bounding boxes (see Figure \ref{fig:t3}). In the worst case, the TOOD's result was evaluated as having two FPs and one FN because the IoU was below 0.5, even though the two bounding boxes accurately marked the fire in the image. This illustrates the potential for using a different method of evaluating bounding boxes in the training process for objects with ambiguous boundaries, which would simplify manual labeling.

\begin{figure}
\centering
\includegraphics[width=\columnwidth]{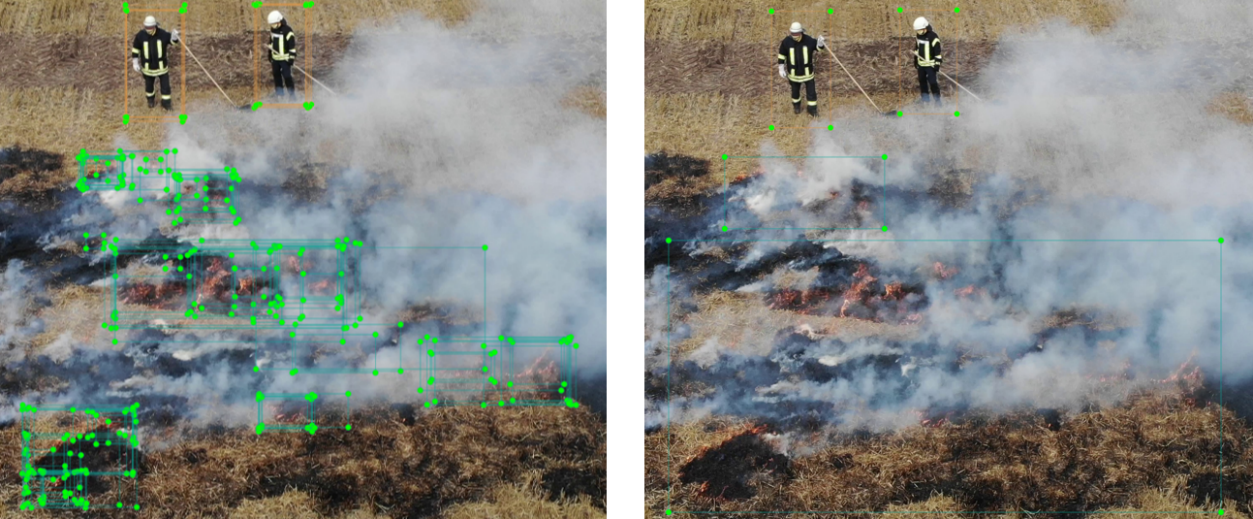}
\caption{Input image (left), filtering and fusion of bounding boxes using the ODP (right).}
\label{fig:t4}
\end{figure}

The potential of using the ODP to combine the results of multiple models is demonstrated in Figure \ref{fig:t4}. The left side of the figure shows the results of all models in one image, where there is significant variation in the bounding boxes. On the right side of the figure are the merged results of the ODP. Optimally, the strengths of the different neural network architectures complement each other and the visualization of the results for further operations is simplified for emergency personnel. For example, the bounding boxes for the 'Fire' and 'Human' classes could be taken exclusively from the TOOD and the bounding boxes for the 'Vehicle' class could be taken from the Deformable DETR. Another possibility is to consider a bounding box to be a true positive only if it is detected by at least three models. This would likely reduce the number of false positives. These changes could be implemented in the ODP with minimal effort.

\section{BUILDING A DATASET}

In this paper, we present a method for creating a new and very important dataset in the COCO format for object detection and classification tasks. The main part of the data set consists of UAV images. We have collected the images over the last 5 years during the missions of the DRZ's Robotics Task Force during various real missions and exercises with the rescue forces. Such images are not publicly available for privacy reasons. To create the dataset, we first ran each of four neural networks (TOOD, YOLOX, VfNet, and Deformable DETR) on all of the images and kept only those bounding boxes that had a result of at least 50\% confidence. This step served to filter out low-confidence detections that may not be reliable. Next, we compared the bounding boxes from the different networks using the intersection over union measure. If at least two networks detected the same class in the same location with at least 50\% confidence, we counted it as a true positive and added the bounding box to the new dataset. This step helped to ensure that the bounding boxes included in the dataset were relatively robust and consistent across different networks. Finally, we annotated the images in the new dataset using the COCO format, which includes information such as class labels and bounding boxes for each object in the image. This allowed us to organize the data in a standardized and structured way that is compatible with a wide range of computer vision tasks.

Overall, our method for creating a new dataset in the COCO format involved a combination of automatic processing. While there is always a risk of errors in any automated process, we believe that the combination of multiple neural networks and strict inclusion criteria helped to improve the quality and reliability of the dataset. After completing the process, we published the dataset on Kaggle, a popular platform for data science competitions and projects, to make it widely available to researchers and practitioners working in the field of computer vision.

\section{CONCLUSIONS}
This paper presents our research using UAVs and computer vision techniques to assist in rescue operations, specifically vegetation fires. Data was acquired from various sources, including simulated images from a 3D game engine. An object detection pipeline was created to label the dataset, employing three pre-trained neural networks, a rule-based algorithm, and bounding box correction filters. Supervised learning was conducted using five MMDetection library models.

Initial learning cycle results indicated potential in object recognition, though with varying strengths and weaknesses among models. To enhance results, false positives were extracted and used in a data augmentation pipeline to create mosaic-like images, tagged using the DAP and added to the dataset. The optimized dataset and second learning run increased false positives but decreased false negatives, improving mean average precision and recall. Models were also able to detect objects outside the original dataset. Filtering and fusing results via the object detection pipeline simplifies presentation to responders. Further improvement could be achieved by targeting specific object classes and weighting neural network results. This approach is explored in the German Rescue Robotics Center (A-DRZ) joint research project\cite{9597869} \cite{9738529}.

\section*{Acknowledgment}
This work was funded by the Federal Ministry of Education and Research (BMBF) under grant number 13N14860 (A-DRZ \url{https://rettungsrobotik.de/}. Thanks to all of our partners in the A-DRZ project).

\bibliographystyle{IEEEtran} 
\bibliography{references_papers} 

\end{document}